\documentclass[conference]{IEEEtran}
\IEEEoverridecommandlockouts
\usepackage{amsmath,amsfonts}
\usepackage{algorithmic}
\usepackage{algorithm}

\usepackage{array}
\usepackage{breqn}
\usepackage[caption=false,font=normalsize,labelfont=sf,textfont=sf]{subfig}
\usepackage{textcomp}
\usepackage{cite}
\usepackage{hyperref}
\hypersetup{
    colorlinks = true,
    allcolors = blue
}

\usepackage{multirow}
\usepackage{stfloats}
\renewcommand{\arraystretch}{1.5} 
\usepackage{url}
\usepackage{pgfplots}
\usepackage{verbatim}
\usepackage{pgfplotstable}
\usepackage{graphicx}
\usepackage{tikz}
\usepackage{color,soul} 
\usepackage{xcolor}
\def\BibTeX{{\rm B\kern-.05em{\sc i\kern-.025em b}\kern-.08em
    T\kern-.1667em\lower.7ex\hbox{E}\kern-.125emX}}
\begin{document}

\title{\large{\textcolor{blue}{\bf This paper was accepted to IEEE MWSCAS 2026}}\\ \LARGE \bf Dual Attention Heads for Personalized Federated Learning in ECG Classification}\vspace{-0.5cm}


\author{\IEEEauthorblockN{Kien Le}
\IEEEauthorblockA{\textit{Computer Science} \\\textit{College of Arts and Sciences}\\
\textit{Florida State University}\\
Tallahassee, FL 32310, USA \\
kl23a@fsu.edu}
\and
\IEEEauthorblockN{Joseph Lindley}
\IEEEauthorblockA{\textit{Chemical \& Biomedical Engineering}\\\textit{FAMU-FSU College of Engineering} \\
\textit{Florida State University}\\
Tallahassee, FL 32310, USA\\
jpl24f@fsu.edu}
\and
\IEEEauthorblockN{Quoc Bao Phan and Tuy Tan Nguyen*}
\IEEEauthorblockA{\textit{Electrical \& Computer Engineering}\\\textit{FAMU-FSU College of Engineering} \\
\textit{Florida State University}\\
Tallahassee, FL 32310, USA \\
qp25c@fsu.edu, tuy.nguyen@fsu.edu}
}

\maketitle

\begin{abstract}

Federated learning (FL) enables collaborative model training across institutions without sharing sensitive patient data. However, the inherent heterogeneity of electrocardiogram (ECG) data across healthcare providers presents significant technical challenges for robust classification. We propose FedDualAtt, a personalized federated learning approach that splits transformer attention heads into global and local branches. Global heads are aggregated via FedAvg to capture shared cross-site patterns, while local heads remain client-specific to adapt to institution-level recording characteristics. Experiments on FedCVD, an FL benchmark for cardiovascular disease detection, demonstrate that FedDualAtt outperforms existing FL and personalized FL methods in ECG classification tasks. Analysis of global-local head ratios reveals that different clients benefit from varying levels of architectural personalization.
\end{abstract}


\section{Introduction}
Cardiovascular diseases account for approximately 19.8 million deaths annually, making them the leading cause of global mortality~\cite{who_cvd_2025}. Although the 12-lead electrocardiogram (ECG) remains the primary non-invasive tool for detecting cardiac pathologies, errors in manual ECG analysis can lead to misdiagnosis and delayed treatment~\cite{sattar_electrocardiogram_2023}. Deep learning models, particularly transformer-based architectures, have shown remarkable performance in automated multi-label ECG classification~\cite{natarajan2020wide}, yet their deployment requires large, diverse training sets drawn from multiple clinical sites to ensure generalizability. 

Federated learning (FL)~\cite{DBLP:journals/corr/McMahanMRA16} addresses these challenges by enabling collaborative model training across healthcare providers without centralizing sensitive patient records. Nevertheless, ECG data exhibits significant heterogeneity across healthcare providers due to variations in recording equipment, patient demographics, and local disease prevalence~\cite{fedcvd}. These non-independently and identically distributed (non-IID) data distributions cause performance degradation in standard FL approaches like FedAvg~\cite{zhao2018federated}, which attempt to enforce a single global model across divergent client distributions. While algorithm-level personalized FL methods, such as Ditto~\cite{ditto} and FedALA~\cite{fedala}, mitigate this through local fine-tuning or adaptive aggregation, they introduce additional training objectives without addressing heterogeneity at the representation level. 

We observe that transformer self-attention is particularly distribution-sensitive: the patterns a head attends to naturally reflect the statistics of its training data. This motivates an architectural personalization strategy that partitions attention heads into two functional groups. Global heads are aggregated via FedAvg to capture universal temporal ECG patterns, while local heads remain client-specific to adapt to site-specific recording characteristics. This partitioning introduces no additional training objectives or hyperparameters beyond the head-split ratio.

Our work introduces the FedDualAtt framework with the following contributions:
\begin{itemize} 
    \item We introduce a dual-attention transformer module appended to the FedCVD~\cite{fedcvd} ResNet1D-34 backbone that partitions attention heads into a global branch (FedAvg-aggregated) and a local branch (per-client), enabling simultaneous cross-site generalization and site-specific adaptation within a single forward pass; 
    \item We design a federated training protocol with strict parameter separation: global and local parameters are stored, transmitted, and aggregated independently; 
    \item We conduct empirical analysis on the FedCVD benchmark over all nine head-ratio configurations, characterizing the stability-performance tradeoff in the global-local attention split. 
\end{itemize} 
The remaining sections are structured as follows: Section~\ref{section II} reviews related work, Section~\ref{section III} presents the proposed method, Section~\ref{section IV} describes experimental evaluation, and Section~\ref{section V} concludes.

\section{Background}
\label{section II}
\subsection{Federated learning for ECG} The FedCVD benchmark~\cite{fedcvd} establishes a standardized evaluation for FL methods on multi-center ECG classification using four real-world datasets with 20 diagnostic labels. Evaluating seven FL algorithms, the best-reported result was Scaffold~\cite{scaffold} at 70.1\% Global Micro-F1. All evaluated methods operate on a ResNet1D-34 backbone without a temporal attention component, leaving sequential ECG dependencies to convolutional layers alone. Our work addresses this gap with FedDualAtt by augmenting the ResNet1D-34 backbone with dual-attention transformer blocks designed to disentangle global and local temporal representations.

\subsection{Personalized federated learning} Parameter-level personalization includes Ditto (dual-objective local fine-tuning with a proximal term), FedALA (local adaptive aggregation weights), FedBN~\cite{fedbn} (per-client batch normalization statistics), and FedProx~\cite{fedprox} (proximal regularization to the global model). These approaches treat personalization as a post-aggregation adaptation step and can be applied to any architecture. FedDualAtt instead enforces the global/local boundary at design time: the parameter partition is fixed at construction, and the FL protocol applies standard FedAvg to the global partition without any additional objectives or gradient manipulation. Head specialization has precedent in natural language processing, where analyses show that individual transformer attention heads learn distinct syntactic and semantic functions~\cite{voita2019analyzing}. FedDualAtt builds on this by making the specialization architecturally explicit, assigning heads to global or local partitions at design time.

\section{Proposed Method}
\label{section III}
\begin{figure*}[!t]
    \centering
    \includegraphics[width=\textwidth]{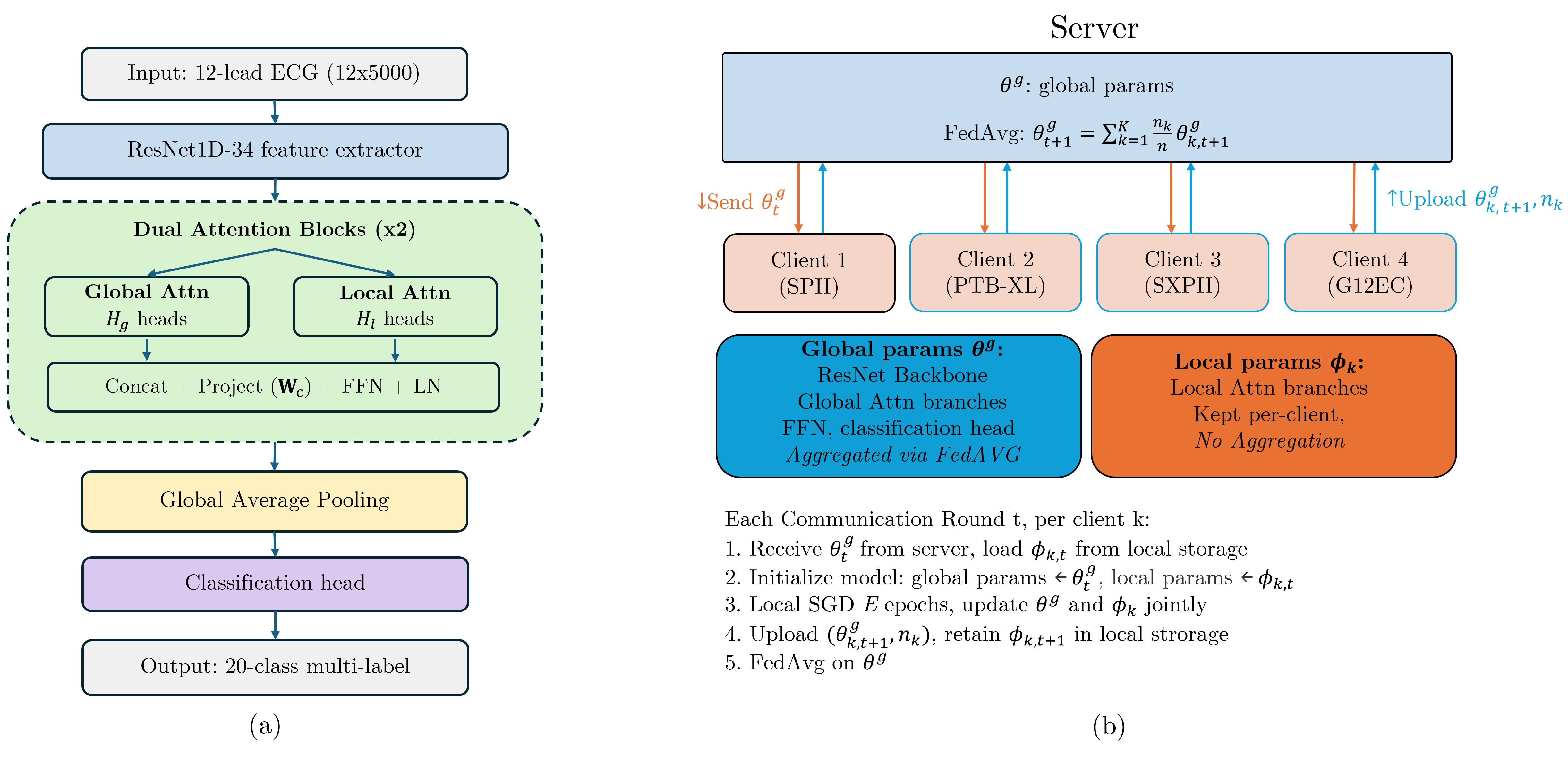}
    \caption{FedDualAtt framework. (a)~DualAttentionResNet1D with parallel global ($\theta^g$, FedAvg-aggregated) and local ($\phi_k$, per-client) attention branches. (b)~Federated training protocol with separate parameter stores for $\theta^g$ and $\{\phi_k\}_{k=1}^K$.}
    \label{fig:framework}
\end{figure*}
We consider a $K = 4$ healthcare providers, each with a private dataset $D_k$ of 12-lead ECG recordings $x^k_i$ and 20-class multi-label binary targets $y^k_i$. Clients collaborate for $T = 50$ communication rounds. 

\subsection{Model Architecture}
Fig.~\ref{fig:framework} illustrates our proposed architecture. We partition the model parameters into two sets: global parameters $\theta^g$ (aggregated across clients) and per-client local parameters $\phi_k$ (never aggregated).

We integrate dual attention into a hybrid CNN-Transformer architecture, with a ResNet1D-34 that extracts features from 12-lead ECG signals, followed by two stacked dual attention transformer blocks, and a classification head for multi-label prediction.

\textbf{ResNet1D-34 feature extractor:}
We adopt the ResNet1D-34 convolutional feature extractor from FedCVD without modification, ensuring direct comparability with all baseline methods. It maps each input $x_i^k \in \mathbb{R}^{12 \times L}$ to a feature sequence $\mathbf{X} \in \mathbb{R}^{L' \times d}$ ($d=512$, $L'= 156$) via strided convolutions, pooling, and sinusoidal positional encoding.

\textbf{Dual attention transformer block:}
The core design insight is that full FedAvg destroys site-specific attention patterns, while purely local training cannot exploit cross-site data. We resolve this by splitting the $H\!=\!8$ attention heads into $H_g$ global heads and $H_l$ local heads ($H_g + H_l = H$): the global heads are aggregated via FedAvg and learn patterns transferable across healthcare providers, while the local heads are kept per-client and adapt to each site's own distribution. Each block processes $\mathbf{X}$ through these two parallel multi-head attention (MHA) branches with layer normalization (LN) and fixed head dimension $d_h=64$. Each branch uses an input projection $\mathbf{W}_{in}$ and output projection $\mathbf{W}_{out}$ to map between model dimension $d$ and the branch's attention space.

\emph{Global branch}:
\begin{equation}
  \hat{\mathbf{G}} =
  \mathrm{LN}\!\left(\mathbf{X} +
  \mathrm{MHA}\!\left(\mathbf{X}\mathbf{W}_{in}^g,\,H_g\right)
  \mathbf{W}_{out}^g\right) \;\in \mathbb{R}^{L' \times d}
  \label{eq:global_branch}
\end{equation}

\emph{Local branch}:
\begin{equation}
  \hat{\mathbf{L}} =
  \mathrm{LN}\!\left(\mathbf{X} +
  \mathrm{MHA}\!\left(\mathbf{X}\mathbf{W}_{in}^l,\,H_l\right)
  \mathbf{W}_{out}^l\right) \;\in \mathbb{R}^{L' \times d}
  \label{eq:local_branch}
\end{equation}

The branches are concatenated and projected to dimension $d$,
then refined through a feed-forward network (FFN):
\begin{align}
  \mathbf{X}' &=
    [\hat{\mathbf{G}};\hat{\mathbf{L}}]
    \mathbf{W}_c \label{eq:combine} \\
  \mathbf{Y} &=
    \mathrm{LN}\!\left(\mathbf{X}' +
    \mathrm{FFN}(\mathbf{X}')\right)
\end{align}
Intuitively, the global branch learns which ECG time steps are mutually relevant across all participating healthcare providers, and because its parameters are FedAvg-aggregated, it is able to capture universal patterns to the entire client population. The local branch performs the same operation with institution-specific parameters $\phi_k$ by learning which temporal patterns are relevant for the patient population of the client $k$ and the recording conditions, and is never shared with other clients. The combine step fuses both representations into a single sequence, and the FFN applies a position-wise nonlinear transformation to each time step independently. The combined projection $\mathbf{W}_c$, FFN, and all three LN layers are global parameters $\theta^g$. Keeping $\mathbf{W}_c$ global ensures that global and local representations are fused in a consistent coordinate space across clients as a per-client $\mathbf{W}_c$ would allow each site to reinterpret the shared global features arbitrarily, thus undermining cross-client alignment.

\textbf{Classification head:}
Global average pooling reduces the sequence dimension of $\mathbf{Y}$, and a fully-connected layer with sigmoid activation produces 20-class multi-label predictions trained with binary cross-entropy loss.

\subsection{Federated Training Protocol}
The server maintains global parameters $\theta^g$ (global attention heads, combine layer, FFN, and classification head). Each client $k$ independently stores its own local parameters $\phi_k$ (local attention heads) without sharing them with the server or other clients.
Algorithm~\ref{alg:feddualatt} describes one communication round of the protocol.
Since $\phi_k$ is never aggregated, each client's local attention heads adapt exclusively to its own data distribution while the shared $\theta^g$ benefits from all clients via FedAvg.
\begin{algorithm}[t]
\caption{FedDualAtt: Communication Round $t$}
\label{alg:feddualatt}
\begin{algorithmic}[1]
\REQUIRE Server: $\theta^g_t$; Client $k$: $\phi_{k,t}$, $D_k$, epochs $E$
\ENSURE Server: updated $\theta^g_{t+1}$; Client $k$: updated $\phi_{k,t+1}$
\FOR{each client $k = 1, \ldots, K$ \textbf{in parallel}}
    \STATE \textbf{Downlink}: receive $\theta^g_t$ from server
    \STATE Initialize model: load $\theta^g_t$ for global params, $\phi_{k,t}$ for local attention
    \STATE \textbf{Training}: run SGD for $E$ epochs on $D_k$, updating all params jointly
    \STATE \textbf{Uplink}: send $(\theta^g_{k,t+1},\; n_k)$ to server
    \STATE Retain $\phi_{k,t+1}$ in local storage
\ENDFOR
\STATE \textbf{Aggregation}: $\theta^g_{t+1} \leftarrow \sum_{k=1}^K \frac{n_k}{n}\,\theta^g_{k,t+1}$, \quad $n = \sum_{k=1}^K n_k$
\RETURN $\theta^g_{t+1}$
\end{algorithmic}
\end{algorithm}

\section{Experimental Evaluation}
\label{section IV}

\subsection{Setup}
Our proposed framework is evaluated using the FedCVD multi-center ECG classification benchmark. This benchmark includes four distinct clinical datasets: Shandong Provincial Hospital (SPH), Physikalisch-Technische Bundesanstalt (PTB-XL), Shaoxing People's Hospital (SXPH), and the PhysioNet 2020 Challenge (G12EC). The four sites differ substantially in dataset size, recording hardware, and diagnostic label prevalence, creating significant non-IID conditions~\cite{fedcvd}. The task is 20-class multi-label ECG classification. We report per-client Micro-F1 and mean average precision (mAP), which measure site-level adaptation, and global Micro-F1 and mAP, which aggregate predictions over all test samples across all four sites and reflect the ability to generalize to new healthcare providers. We compare FedDualAtt against standard federated algorithms, including FedAvg, FedProx, and Scaffold, as well as personalized methods such as Ditto and FedALA. For FedDualAtt, we evaluated across all nine head-ratio configurations ($8H_g\!:\!0H_l$ to $0H_g\!:\!8H_l$). All FedDualAtt results are mean$\pm$std over 5 seeds.
Baseline FL results (Micro-F1, mAP) are taken from the FedCVD paper.

\subsection{Main Results}

Table~\ref{tab:main} reports per-client and global Micro-F1 and mAP for all methods.
Among all configurations, FedDualAtt with $8H_g\!:\!0H_l$ (all global heads) achieves the highest Global Micro-F1 at \textbf{72.7\%}, surpassing the previous best (Scaffold, 70.1\%) by $2.6$ percentage points (pp).
This improvement comes entirely from the transformer attention architecture augmenting the ResNet1D-34 backbone, not from personalization. The local-only extreme ($0H_g\!:\!8H_l$) achieves competitive per-client F1 but collapses global Micro-F1 to 50.8\%, confirming that cross-client aggregation is essential for generalization across sites.

Introducing local heads (any ratio with $H_l \geq 1$) consistently improves per-client Micro-F1: SPH reaches 86.6--87.8\% (vs.\ FedALA 84.4\%), PTB-XL reaches 70.1--75.2\% (vs.\ FedALA 71.7\%), and G12EC reaches 70.7--74.3\% (vs.\ Ditto 73.4\%).
The exception is SXPH, where FedALA (88.2\%) remains the strongest, suggesting that adaptive aggregation is more effective for that distribution.
Notably, Ditto shows high variance on G12EC ($\pm$6.7 F1), a known failure mode of proximal-term methods on highly heterogeneous clients. FedDualAtt remains stable across all four sites.

\begin{table*}[!t]
  \renewcommand{\arraystretch}{1.35}
  \centering
  \caption{ECG benchmark: per-client and global Micro-F1 / mAP (\%).
    Baseline FL rows report mean$\pm$std from the FedCVD paper.
    FedDualAtt rows report mean$\pm$std over 5 seeds.
    \textbf{Bold} = best, \underline{underline} = second-best, ranked jointly.}
  \label{tab:main}
  \begin{tabular}{|c|l|cc|cc|cc|cc|cc|}
    \hline
    \multirow{2}{*}{} & \multirow{2}{*}{Method}
      & \multicolumn{2}{c|}{SPH}
      & \multicolumn{2}{c|}{PTB-XL}
      & \multicolumn{2}{c|}{SXPH}
      & \multicolumn{2}{c|}{G12EC}
      & \multicolumn{2}{c|}{Global} \\
    \cline{3-12}
    & & F1 & mAP & F1 & mAP & F1 & mAP & F1 & mAP & F1 & mAP \\
    \hline
    \multirow{5}{*}{FedCVD} & FedAvg  & 69.0{\scriptsize$\pm$10.1} & 58.5{\scriptsize$\pm$1.2} & 50.3{\scriptsize$\pm$5.3} & 54.4{\scriptsize$\pm$0.5} & 77.6{\scriptsize$\pm$0.7} & 37.2{\scriptsize$\pm$0.3} & 66.3{\scriptsize$\pm$0.9} & 39.5{\scriptsize$\pm$0.5} & 67.9{\scriptsize$\pm$3.8} & 50.8{\scriptsize$\pm$0.4} \\
    & FedProx & 74.0{\scriptsize$\pm$7.5} & 60.3{\scriptsize$\pm$2.9} & 55.6{\scriptsize$\pm$2.7} & 56.4{\scriptsize$\pm$0.6} & 73.2{\scriptsize$\pm$1.0} & 36.0{\scriptsize$\pm$0.8} & 70.2{\scriptsize$\pm$2.3} & \underline{43.8{\scriptsize$\pm$1.8}} & 68.8{\scriptsize$\pm$2.6} & \underline{52.3{\scriptsize$\pm$0.9}} \\
    & Scaffold & 77.5{\scriptsize$\pm$2.6} & 58.0{\scriptsize$\pm$1.2} & 56.9{\scriptsize$\pm$1.7} & 55.9{\scriptsize$\pm$0.7} & 73.3{\scriptsize$\pm$1.0} & 36.2{\scriptsize$\pm$0.6} & 70.7{\scriptsize$\pm$2.9} & 42.7{\scriptsize$\pm$1.1} & \underline{70.1{\scriptsize$\pm$0.8}} & 52.1{\scriptsize$\pm$0.7} \\
    & Ditto   & 82.8{\scriptsize$\pm$4.4} & \textbf{63.1{\scriptsize$\pm$4.2}} & \underline{74.8{\scriptsize$\pm$1.4}} & 58.3{\scriptsize$\pm$0.6} & 86.5{\scriptsize$\pm$1.5} & 38.1{\scriptsize$\pm$0.6} & 73.4{\scriptsize$\pm$6.7} & 42.2{\scriptsize$\pm$4.0} & 68.1{\scriptsize$\pm$2.9} & 48.7{\scriptsize$\pm$1.4} \\
    & FedALA  & 84.4{\scriptsize$\pm$4.0} & \underline{62.0{\scriptsize$\pm$7.0}} & 71.7{\scriptsize$\pm$5.7} & 57.1{\scriptsize$\pm$2.2} & \textbf{88.2{\scriptsize$\pm$0.1}} & 37.4{\scriptsize$\pm$0.2} & 66.7{\scriptsize$\pm$5.9} & 41.2{\scriptsize$\pm$2.3} & 67.8{\scriptsize$\pm$1.9} & 50.8{\scriptsize$\pm$1.3} \\
    \hline
    \multirow{9}{*}{FedDualAtt} & $8H_g\!:\!0H_l$ & 80.7{\scriptsize$\pm$2.2} & 57.3{\scriptsize$\pm$1.0} & 56.7{\scriptsize$\pm$2.1} & 56.5{\scriptsize$\pm$2.6} & 78.1{\scriptsize$\pm$1.2} & 39.7{\scriptsize$\pm$1.0} & 68.3{\scriptsize$\pm$1.5} & 42.6{\scriptsize$\pm$1.9} & \textbf{72.7{\scriptsize$\pm$1.1}} & \textbf{54.8{\scriptsize$\pm$1.9}} \\
    & $7H_g\!:\!1H_l$ & \underline{87.5{\scriptsize$\pm$0.2}} & 59.4{\scriptsize$\pm$0.6} & 72.3{\scriptsize$\pm$0.7} & 59.6{\scriptsize$\pm$0.9} & \underline{86.8{\scriptsize$\pm$0.6}} & 40.2{\scriptsize$\pm$0.5} & 73.6{\scriptsize$\pm$0.9} & 43.6{\scriptsize$\pm$0.5} & 63.4{\scriptsize$\pm$3.0} & 46.0{\scriptsize$\pm$0.4} \\
    & $6H_g\!:\!2H_l$ & 87.0{\scriptsize$\pm$1.1} & 56.6{\scriptsize$\pm$3.8} & 70.1{\scriptsize$\pm$4.1} & 57.7{\scriptsize$\pm$4.7} & 85.7{\scriptsize$\pm$2.6} & 38.4{\scriptsize$\pm$2.2} & 70.7{\scriptsize$\pm$3.4} & 40.7{\scriptsize$\pm$2.6} & 57.5{\scriptsize$\pm$2.8} & 43.8{\scriptsize$\pm$4.1} \\
    & $5H_g\!:\!3H_l$ & \textbf{87.8{\scriptsize$\pm$0.5}} & 59.2{\scriptsize$\pm$0.6} & 74.0{\scriptsize$\pm$1.1} & \textbf{61.3{\scriptsize$\pm$1.2}} & 86.2{\scriptsize$\pm$0.4} & \underline{40.5{\scriptsize$\pm$0.4}} & 73.2{\scriptsize$\pm$1.7} & 43.1{\scriptsize$\pm$1.0} & 61.9{\scriptsize$\pm$4.9} & 48.4{\scriptsize$\pm$2.2} \\
    & $4H_g\!:\!4H_l$ & 87.0{\scriptsize$\pm$0.6} & 58.8{\scriptsize$\pm$2.2} & 72.6{\scriptsize$\pm$1.7} & 59.3{\scriptsize$\pm$2.4} & 85.7{\scriptsize$\pm$1.2} & 40.1{\scriptsize$\pm$1.4} & 73.2{\scriptsize$\pm$2.3} & 43.3{\scriptsize$\pm$2.1} & 65.6{\scriptsize$\pm$3.3} & 50.3{\scriptsize$\pm$2.4} \\
    & $3H_g\!:\!5H_l$ & 87.4{\scriptsize$\pm$0.6} & 59.7{\scriptsize$\pm$0.9} & 73.1{\scriptsize$\pm$1.1} & 60.5{\scriptsize$\pm$1.0} & 85.7{\scriptsize$\pm$1.1} & \textbf{40.5{\scriptsize$\pm$0.4}} & 73.7{\scriptsize$\pm$1.9} & \textbf{43.8{\scriptsize$\pm$1.3}} & 65.1{\scriptsize$\pm$6.2} & 51.0{\scriptsize$\pm$4.2} \\
    & $2H_g\!:\!6H_l$ & 86.8{\scriptsize$\pm$0.9} & 56.4{\scriptsize$\pm$4.0} & 70.3{\scriptsize$\pm$3.5} & 57.4{\scriptsize$\pm$3.6} & 85.2{\scriptsize$\pm$1.7} & 38.6{\scriptsize$\pm$2.6} & 71.2{\scriptsize$\pm$3.8} & 41.7{\scriptsize$\pm$2.9} & 63.4{\scriptsize$\pm$5.9} & 49.9{\scriptsize$\pm$3.5} \\
    & $1H_g\!:\!7H_l$ & 86.6{\scriptsize$\pm$0.5} & 59.3{\scriptsize$\pm$0.9} & 72.2{\scriptsize$\pm$0.5} & 60.2{\scriptsize$\pm$1.1} & 85.2{\scriptsize$\pm$0.6} & \textbf{40.6{\scriptsize$\pm$0.4}} & \textbf{74.3{\scriptsize$\pm$1.0}} & 43.4{\scriptsize$\pm$0.4} & 66.8{\scriptsize$\pm$5.5} & 51.1{\scriptsize$\pm$1.9} \\
    & $0H_g\!:\!8H_l$ & 87.5{\scriptsize$\pm$0.4} & 59.7{\scriptsize$\pm$1.9} & \textbf{75.2{\scriptsize$\pm$1.8}} & \underline{61.1{\scriptsize$\pm$2.6}} & 86.4{\scriptsize$\pm$1.2} & 39.9{\scriptsize$\pm$1.3} & \underline{74.1{\scriptsize$\pm$3.0}} & 43.8{\scriptsize$\pm$2.7} & 50.8{\scriptsize$\pm$1.7} & 44.5{\scriptsize$\pm$2.4} \\
    \hline
  \end{tabular}
\end{table*}

\subsection{Head-Ratio Ablation}

\begin{figure}[!t]
    \centering
    \includegraphics[width=\columnwidth]{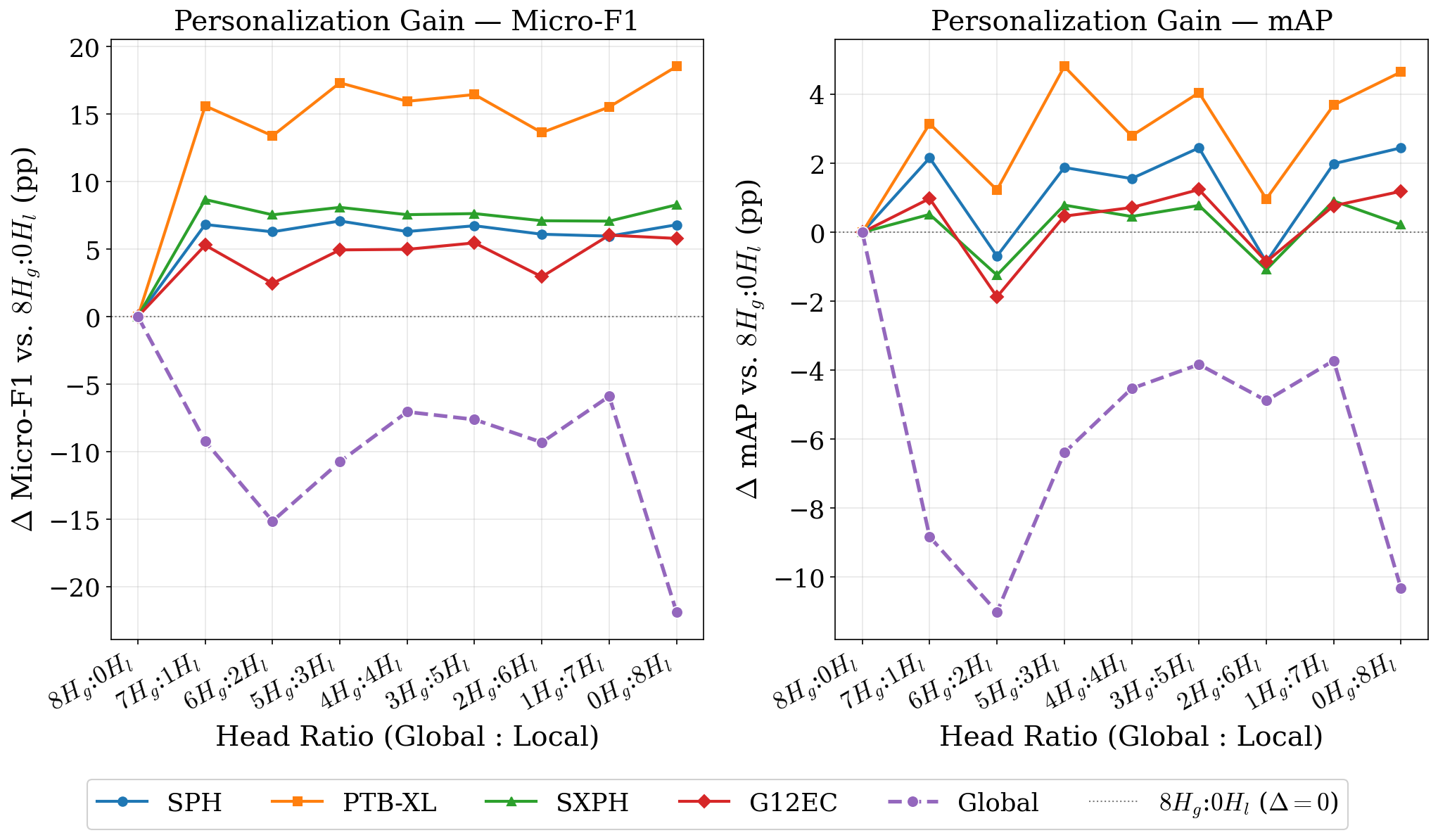}
    \caption{Per-client and global Micro-F1 (left) and mAP (right) delta relative to
    FedDualAtt $8H_g\!:\!0H_l$ across all head-ratio configurations ($n\!=\!5$ seeds).
    Each colored line is one client; the dashed purple line is the global metric.
    Positive values indicate gain over the global-only baseline.}
    \label{fig:delta}
\end{figure}

Fig.~\ref{fig:delta} shows per-client and global Micro-F1 and mAP \emph{delta} relative to the $8H_g\!:\!0H_l$ global-only baseline, with one line per client across all nine head-ratio configurations.
We can see three patterns emerge.
First, introducing even a single local head ($7H_g\!:\!1H_l$) produces an immediate and substantial per-client Micro-F1 gain for most clients ($+5$ to $+10$ pp on SPH, PTB-XL, and G12EC), confirming that local attention heads rapidly capture site-specific patterns that global aggregation suppresses.
Second, per-client gains are largely stable from $7H_g\!:\!1H_l$ onward, with the per-client curves plateauing across the middle configurations ($6H_g\!:\!2H_l$ through $1H_g\!:\!7H_l$), indicating that a small number of local heads is sufficient to capture most of the personalization benefit.
Third, global Micro-F1 (dashed purple) declines monotonically as local heads increase, dropping $-21.9$ pp at $0H_g\!:\!8H_l$ relative to $8H_g\!:\!0H_l$, because fully local attention cannot leverage cross-client information during aggregation.
The same global--local trade-off is visible in the mAP panel, with per-client mAP also improving under moderate personalization while global mAP falls at high local-head counts.
Taken together, the head ratio $H_g\!:\!H_l$ provides an interpretable knob for controlling this trade-off, requiring no additional training objectives or hyperparameters.

\section{Conclusion}
\label{section V}
In this paper, we propose FedDualAtt, which addresses data heterogeneity in federated ECG classification by splitting transformer attention heads into a globally aggregated branch and a per-client local branch.
Experiments on the FedCVD benchmark demonstrate two complementary benefits: the dual-attention architecture alone ($8H_g\!:\!0H_l$) raises Global Micro-F1 to 72.68\%, surpassing all FL baselines, while introducing local heads consistently improves per-client performance on three of four clients with gains that plateau after a single local head.
Our ablation reveals a monotonic trade-off between global generalization and per-client adaptation that is controlled entirely by the head ratio $H_g\!:\!H_l$, without additional objectives or training phases. These results suggest that architectural personalization at the attention level is an effective complement to algorithm-level FL methods for heterogeneous clinical data. The dual-head partitioning principle extends naturally to other federated medical domains where clients share backbone features but differ in local signal characteristics, such as multi-site electroencephalography-based seizure detection or distributed radiology. A promising direction for future work is automatic ratio selection, where each client learns a soft assignment of heads to the global or local pool, eliminating the need to manually tune $H_g\!:\!H_l$.

\section*{Acknowledgment}
This research was supported by the National Science Foundation under the Directorate for Computer and Information Science and Engineering/Office of
Advanced Cyberinfrastructure (CISE/OAC) Grant No. 2600417, and was partially supported by the Florida State University Undergraduate Research Opportunity Program -- Research Mentor Materials Grant.

\bibliographystyle{IEEEtran}
\bibliography{refs}

\end{document}